\documentclass[conference]{IEEEtran}
\IEEEoverridecommandlockouts
\usepackage{cite}
\usepackage{amsmath,amssymb,amsfonts}
\usepackage{algorithmic}
\usepackage{graphicx}
\usepackage{textcomp}
\usepackage{xcolor}
\usepackage{hyperref}
\usepackage{amssymb}
\usepackage{booktabs}
\usepackage{multirow}
\usepackage{algorithm}
\usepackage{algorithmic}
\newcommand{\algref}[1]{Algorithm~\ref{alg:#1}}

\newcommand{\figref}[1]{Figure~\ref{fig:#1}}
\newcommand{\tabref}[1]{Table~\ref{tab:#1}}
\newcommand{\equref}[1]{Eq.~\ref{equ:#1}}

\newcommand{\eg}{\textit{e}.\textit{g}.}

\def\BibTeX{{\rm B\kern-.05em{\sc i\kern-.025em b}\kern-.08em
    T\kern-.1667em\lower.7ex\hbox{E}\kern-.125emX}}
\begin{document}

% \author{\IEEEauthorblockN{Anonymous Authors}}
\title{Patch-wise Mixed-Precision Quantization of Vision Transformer
}

\author{\IEEEauthorblockN{Junrui Xiao$^{1,2}$, Zhikai Li$^{1,2}$, Lianwei Yang$^{1,2}$, Qingyi Gu$^{1,*}$\thanks{$^{*}$Corresponding author at: 95 Zhongguancun East Road, Beijing 100190,
China. E-mail addresses: qingyi.gu@ia.ac.cn}}
\IEEEauthorblockA{
\textit{$^1$Institute of Automation, Chinese Academy of Sciences.} Beijing, China\\
\textit{$^2$School of Artificial Intelligence, University of Chinese Academy of Sciences.} Beijing, China\\
\{xiaojunrui2020, lizhikai2020, yanglianwei2021, qingyi.gu\}@ia.ac.cn}
}
\maketitle

\begin{abstract}
As emerging hardware begins to support mixed bit-width arithmetic computation, mixed-precision quantization is widely used to reduce the complexity of neural networks. However, Vision Transformers (ViTs) require complex self-attention computation to guarantee the learning of powerful feature representations, which makes mixed-precision quantization of ViTs still challenging. In this paper, we propose a novel patch-wise mixed-precision quantization (PMQ) for efficient inference of ViTs. Specifically, we design a lightweight global metric, which is faster than existing methods, to measure the sensitivity of each component in ViTs to quantization errors. Moreover, we also introduce a pareto frontier approach to automatically allocate the optimal bit-precision according to the sensitivity. To further reduce the computational complexity of self-attention in inference stage, we propose a patch-wise module to reallocate bit-width of patches in each layer. Extensive experiments on the ImageNet dataset shows that our method greatly reduces the search cost and facilitates the application of mixed-precision quantization to ViTs.

\end{abstract}


\section{Introduction}
\label{sec:intro}
Recently, transformer-based architectures have been widely adopted and achieve state-of-the-art results in many computer vision tasks such as image classification \cite{vit,deit}, object detection \cite{detr, swin}, and semantic segmentation \cite{setr}.
Although vision transformers (ViTs) are pushing the limits of performance across various tasks, the high computation and memory cost severely hinders their deployment on resource-limited devices.
Numerous model compression and acceleration methods have been proposed to tackle this issue, such as pruning, knowledge distillation, quantization, and efficient model design.
Among those methods, quantization has been one of the effective techniques for compressing neural networks.

Most of the existing quantization methods use uniform bit-width assignment, also known as fix-point quantization\cite{FixedPoint,Fixed-point}, which is suboptimal.
As emerging hardware begins to support mixed bit-width inference, mixed precision quantization (MPQ) has been brought into the spotlight.
Generically, MPQ can be represented as a projection $Q:\mathcal{X}\in\mathbb{R}\rightarrow \mathcal{Q}_x\in\{q_0,q_1,...,q_K\}$, which quantifies the real-valued weights and activations to the various lower bit-width integers.
In contrast to the fix-point quantization, MPQ can fully leverage the difference of representative capacity and redundancy in various components of deep neural networks.
Consequently, by finding the optimal bit-width for each component, MPQ can achieve a better trade-off between accuracy and efficiency.
However, the MPQ of ViTs is still challenging for two reasons as follows.

First, different from the mainstream CNNs, the ViTs use patches as input and calculate the features of all these patches in parallel.
Multi-head Self-attention (MHSA) will further aggregate all patch embeddings into visual features as the output. 
The attention map represents the relationship between the different patches, essential for capturing global information flow.
However, the key to retaining the performance of quantitative models is how to preserve this global information flow, which cannot be guaranteed in the conventional MPQ of CNNs.

Second, it is not easy to find an optimal quantitative strategy. Prior methods addressed this issue for MPQ are primarily based on searching or metrics.
Search-based methods leverage reinforcement learning or Neural Architecture Search to determine the optimal bit-width.
However, the exponential search space and the time-consuming calculation make it difficult to be applied to ViTs. 
Specifically, for a network with $N$ layers and $K$ candidate bit-widths in each layer, an iterative search scheme has exponential time complexity $(\mathcal{O}(M^N))$.
Metric-based methods aim to reduce the time cost through kinds of metrics that are easy to compute.
However, existing metrics based on the Hessian matrix are not suitable for ViTs.
For example, there exist 7.2M parameters for each encoder layer in ViT-Base, and it is unstable to compute a Hessian matrix of size $7M\times7M$.

To address the above problems, in this paper, we propose a novel patch-wise mixed-precision quantization (PMQ) to tackle the challenges for efficient inference of ViT.
Concretely, as shown in \figref{overall}, we argue that not all patches in each layer contain sufficient discriminative information. 
Hence, we design an Adaptive Attention Shrink (AAS) module that can quantize discriminate patches and redundant patches with different bit-width to preserve global information flow during the inference phase. 
Moreover, we design an efficient Global Sensitivity Ranking (GSR) criterion with a first-order metric to measure the importance of each layer, which can achieve comparable performance to Hessian-based metric with considerable speedup.

To demonstrate the effectiveness of PMQ, we conduct extensive experiments on various ViTs.
For DeiT-Small on the ImageNet-1K dataset, our post-training quantized model can achieve top-1 accuracy 76.68\% with average bit-width of both weights and activations no more than 6 bits.

The contributions of this paper can be summarized as follows:
\begin{itemize}
  \item We design an AAS module that can identify the importance of each patch, and automatically allocate different bit-width according to their discrimination.
  \item To reduce the time complexity of sensitivity ranking, we propose a GSR criterion that can speed up the ranking procedure with comparable performance. 
  \item Extensive experiments are conducted on various ViTs to demonstrate the effectiveness of our proposed method. The results show that our post-training quantized ViTs can achieve comparable performance.
\end{itemize}

\section{Related works}
\subsection{Network Quantization}
\textbf{Fix-point Quantization.} Fixed-point quantization focuses on using the same bits for different layers of the network, and they are usually divided into two categories:
1) Quantization-Aware Training (QAT) methods\cite{Degree-Quant,QAT2,QAT1,QAT3,QAT4} first complete the quantization of the pre-trained model and then retrain it to fine-tune the parameters. Although the accuracy of this method is considerable, the time-consuming training process and complex hyperparameters lead to limited applications. 
2) Post-Training Quantization (PTQ) methods\cite{BRECQ,Near-Lossless,PTQ1,PTQ2,PTQ3}, instead, only requires a small subset of dataset to calibrate the pre-trained model, and then complete the quantization using the quantization parameters obtained from the calibration.

\noindent\textbf{Mixed-precision Quantization.}  To achieve a better trade-off between accuracy and compression ratio, MPQ has attracted more attention in recent years, which can further exploit the different representations of each layer in networks.
Existing MPQ methods can be generally divided into three types:
1) The Metric-Based Methods design a criterion to measure the sensitivity of each layer and thus determine the bit configuration. 
HAWQ-series \cite{HAWQV1,HAWQV2,HAWQV3} use a hessian-based metric to conduct MPQ strategy. OMPQ \cite{OMPQ} design a criterion by exploiting orthogonality between layers of the network. SAQ \cite{SAQ} ranks the sensitivity with sharpness.
2) The Search-Based Methods search optimal quantization strategies with a certain number of evaluations. 
HAQ \cite{HAQ} and AutoQ\cite{AutoQ} use reinforcement learning to allocate bit-width for each layer. DNAS \cite{DNAS} utilizes Neural Architecture Search (NAS) to achieve a differentiable search process. 
However, these methods suffer from large search space and are time-consuming, which is not suitable for large models.        
3) The Optimization-Based Methods consider allocation as an optimization problem. 
FracBits introduce fractional bit-width parameters in quantization. DQ \cite{DQ} uses additional parameters to enable the forward of quantization differentiable. However, these methods introduce some useless optimization variables and need to retrain the model. 

\subsection{Vision Transformer}
The recent transform-based models \cite{vit,deit,detr,swin,setr} have significantly improved accuracy on a wide range of computer vision tasks.
Generally, ViTs split images into a sequence of flattened patches as input. 
Each block has a multi-head self-attention (MHSA) to extract features with global information flow. 
Although ViTs have demonstrated great potential across various tasks, the large amount of memory, computation, and energy consumption hinders their deployment in resource-limited devices, such as mobile and embedding devices. Thus, compression approaches for ViTs are necessary. Recently, several works focus on lightweight architectural design. MobileViT \cite{mobilevit} attempts to incorporate attention into MobileNetV2 \cite{mobilenetv2} and proposes a MobileViT block to enhance the local-global representation of mobile CNNs.
MiniViT\cite{minivit} multiplexes the weights of continuous transformer blocks to reduce the number of parameters and uses weight distillation to improve the accuracy of visual transformers.
Evo-ViT \cite{evovit} uses a slow-fast updating mechanism to accelerate the training and inference.
Le-ViT \cite{levit} proposes a redesigned block to make the process of inference faster.

These excellent methods have inspired current research. In this paper, we make a further step to explore the \textbf{post-training}, \textbf{metric-based}, \textbf{mixed-precision quantization} for effective inference of ViTs.

\begin{figure*}[t!]
  \centering
  \includegraphics[width=\linewidth]{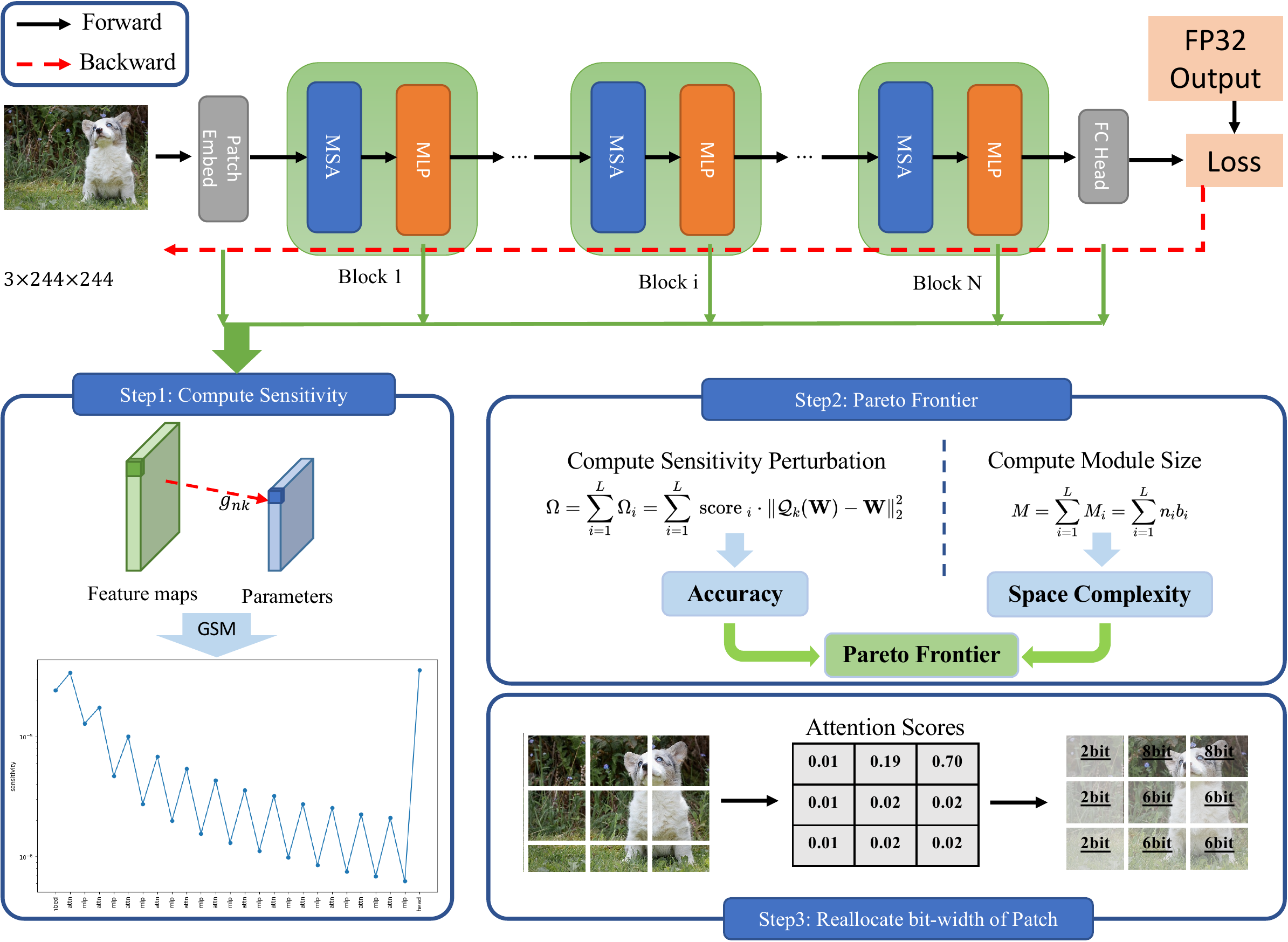}
  \caption{The overview of our proposed PMQ.
  Step 1: Compute the sensitivity by Global Sensitivity Metric to measure the importance of each module. Step 2: Compute Perturbation  and model size to generate Pareto Frontier.
  Step 3: In inference phase, use Adaptive Attention Shrink to identity the importance of each patch, and automatically reallocate different bit-width according to their discrimination.
}
  \label{fig:overall}
\end{figure*}
\section{Method}
In this section, we first present the formulation of ViT quantization, and then introduce the proposed PMQ of vision transformer in detail. The overall pipeline is shown in \figref{overall}.
\subsection{Background}
\textbf{Vision Transformers} commonly consist of multiple transformer encoder layers stacked up together with $N$ patches as input. 
In each basic transformer encoder, there is an MSA followed by an MLP block, which are the main components occupying most of the computational cost.
Denoting the input features to MSA in the $l^{th}$ layer by $X^l\in \mathbb{R}^{N\times d}$, where N is the sequence length and d is the embedding dimension, the MSA can be formulated as:
\begin{equation}
  \text{MSA}(X^l)=\text{Concat}[\text{softmax}(\frac{Q_h^l{K_h^l}^T}{\sqrt{d_h}})V_h^lW_o^l]^H_{h=1},
\end{equation}
where the query, key, and value from $n^{th}$ head in $l^{th}$ layer can be computed by $Q_h^l=X^lW_h^{lQ}$, $K_h^l=X^lW_h^{lK}$, and $V_h^l=X^lW_h^{lV}$ ($W_h^Q,W_h^K,W_h^V\in\mathbb{R}^{d\times d_h}$). 
$H$ is the number of heads and $W_o^l\in\mathbb{R}^{d\times d_h}$ represents the output projection.
The input of MLP is $Z^l=\text{LN}(\text{MSA}(X^l)+X^l)$, and MLP module can be formulated as:
\begin{equation}
  \text{MLP}(Z^l)=\sigma(Z^lW_1^l+b_1^l)W_2^l+b_2^l,
\end{equation}
where $\sigma$ denotes the nonlinear activation function (\eg, GELU), and $W_1,W_2^T\in\mathbb{R}^{d\times d'}$, $b_1\in\mathbb{R}^{d'}$, $b_2\in\mathbb{R}^d$ are projection matrices and biases, respectively.
Thus, the $l^{th}$ transformer encoder block $\text{B}^l(\cdot)$ can be defined as:
\begin{equation}
  \text{B}^l(X^l) = \text{MLP}(Z^l)+Z^l,
\end{equation}
where
\begin{equation}
  Z^l=\text{LN}(\text{MSA}(X^l)+X^l).
\end{equation}

\noindent\textbf{Neural Network Quantization} aims to quantify the real-valued weights and activations ($W, A\in\mathbb{R}$) to the various lower bit-width integers ($\mathcal{Q}_w, \mathcal{Q}_a\in\mathbb{I}$), which significantly decreases the consumption of memory, computation, and data transfer.
Uniform symmetric quantization is the most widely used method, which can be formulated as:
\begin{equation}
  X_q=\text{clamp}(\lfloor\frac{X}{s}\rceil, \textrm{2}^{k-1}, \textrm{2}^{k-1}-\textrm{1}),
\end{equation}
where the $s$ denoting scale and $b$ denoting bit-width are the quantization parameters. 
Post-training quantization (PTQ) leverages a subset of unlabeled images to determine the scaling factors $s$ of activations and weights for each layer.
In this paper, we quantize all the weights and inputs involved in matrix multiplication.

\subsection{Global Sensitivity Metric}
The purpose of MPQ is to reduce computational and memory costs while minimizing task loss by assigning appropriate bit widths to the components of the neural network.
Given a model parameterized by $\textbf{W}=\{w_0,w_1,\cdots,w_M\}$ and a dataset $\mathcal{D}=\{(x_0,y_0),(x_1,y_1),\cdots,(x_N,y_N)\}$, the task of training is to minimize the loss $\mathcal{L}$:
\begin{equation}
  \min _{\mathcal{Q}_k(\mathbf{W})} \mathcal{L}(\mathcal{D}, \mathcal{Q}_k(\mathbf{W})) \quad \text { s.t. } \operatorname{Cost}(\mathrm{ \mathcal{Q}_k(\mathbf{W})}) \leq C,
  \label{equ:task}
\end{equation}
where the $\mathcal{Q}_k(\cdot)$ denotes quantizing $\textbf{W}$ to $k$ bit-width, $\operatorname{Cost}(\cdot)$ denotes the quantization budgets, such as model size and BitOPs.
One way to sovle the \equref{task} is to follow HAWQ \cite{HAWQV2} that use the second-order Taylor expansion of $\mathcal{L}$ and the trace of Hessian as the metric:
\begin{equation}
  \mathcal{L}(w+\Delta w)-\mathcal{L}(w)\approx g_w^T \Delta w+\frac{1}{2} \Delta w^T H_w \Delta w.
\end{equation}

Inspired by \cite{FIM1,FIM2}, we design a lightweight metric to measure the sensitivity of components in ViTs. 
The importance of a parameter can be quantified by the error induced by removing it. Under an i.i.d. assumption, removing the $k_{th}$ parameter (setting $w_k=0$) would lead to:
\begin{equation}
  \mathcal{L}(w-w_km_k)-\mathcal{L}(w)\approx -g_kw_k+\frac{1}{2} H_{kk} w_k^2,
\end{equation}
where $m_k$ is the mask vector, except that the $k{th}$ entry is 1, all other places are zero. $g_k$ and $H_{kk}$ are the gradient and the Hessian matrix, respectively.
Given the pre-trained model is converged to a minimum, we assume that the current parameters $\textbf{W}$ is at a local optimum and the gradient $g_k$ can therefore be thought to be close to 0. 
Since the complexity of the Hessian matrix is quadratic to the number of parameters that makes it much harder to compute especially for ViTs, we approximate the Hessian $H_{kk}$ with the empirical Fisher information matrix $I$:
\begin{equation}
  I=\frac{1}{N} \sum_{i=1}^{N} \nabla \log p\left(x_{i} \mid w\right) \nabla \log p\left(x_{i} \mid w\right)^{\mathrm{T}},
\end{equation}
where the $x_i$ denote the $i_{th}$ element from training data $\mathcal{D}$
Hence, If we use $N$ data points to estimate the Fisher information, the importance of the $i_{th}$ parameter becomes:
\begin{equation}
  score_i=\frac{1}{2 N} w_i^2 \sum_{n=1}^N g_{n i}^2.
  \label{equ:score}
\end{equation}
Since the gradient $g_{nk}$ is already available from backpropagation, the importance score in \equref{score} can be easily calculated without additional cost.

As shown in \figref{overall}, since the MSA always is more sensitive than MLP in the same block, MSA should be quantized to a higher bit-width than MLP, \eg, 2 bits for MLP and 4 bits for MSA.
This is because the attentional map representing the relationship between different patches is crucial to capture the global information flow.
Therefore, we will next explore the impact of different patches to further reduce the computation complexity of MSA.

\subsection{Adaptive Attention Shrink}
In MSA, the relevance between pairs of patches $x_i$ and $x_j$ can be measured by:
\begin{equation}
  \text{Attn}^l_h(x_i,x_j)=\text{softmax}(\frac{x^T{W_h^{lQ}}^TW_h^{lK}}{\sqrt{d_h}}),
\end{equation}
where the $h$ and $l$ denotes head and layer. The computational complexity of the attention matrix ($\mathcal{O}(4d^2n+2n^2d)$) is quadratic with the number of patches.

\begin{figure}[t!]
  \centering
  \includegraphics[width=\linewidth]{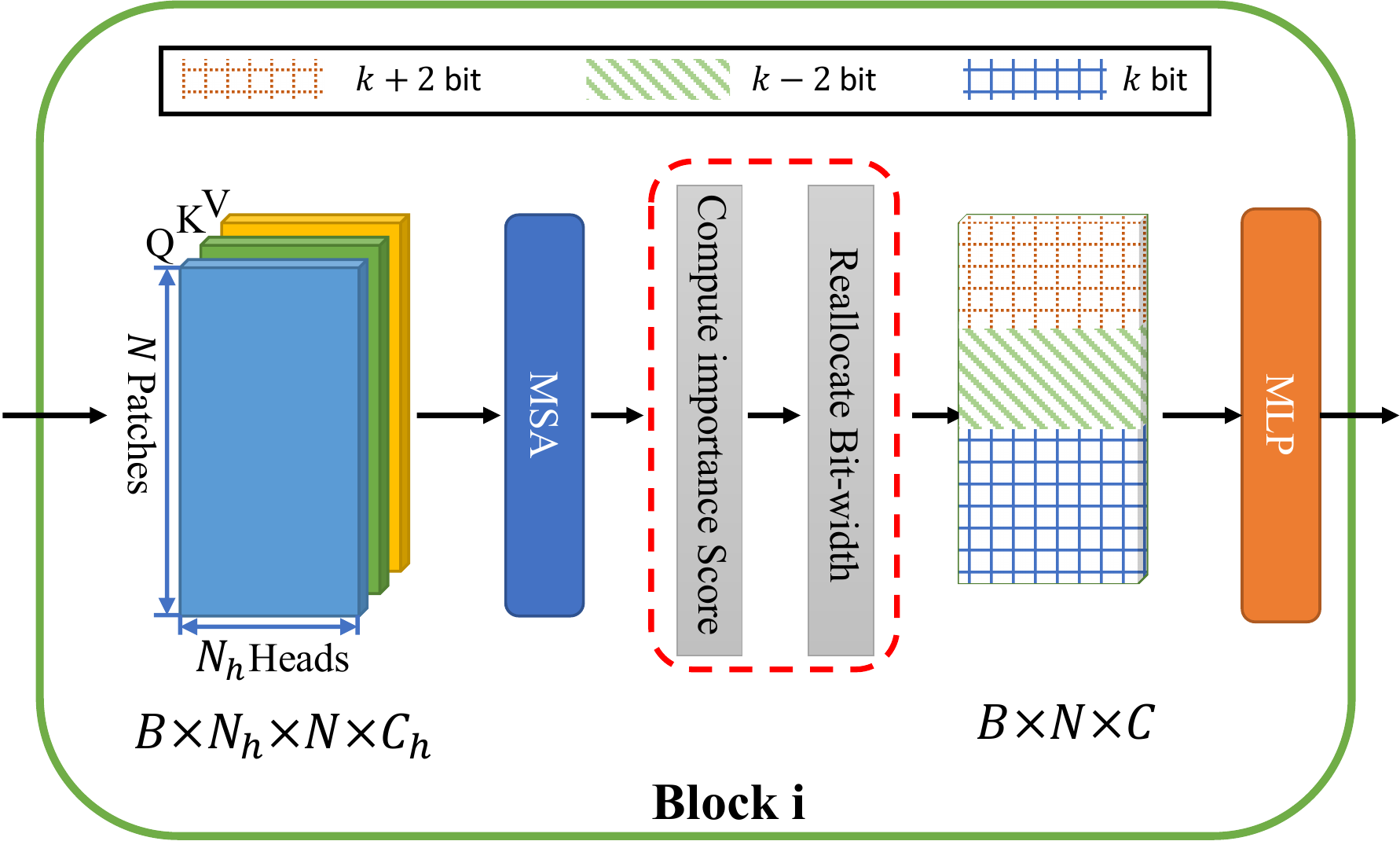}
  \vspace{-0.5cm}
  \caption{Adaptive Attention Shrink: Patches are quantized to different bit-width according to their discrimination in the inference phase.}
  \label{fig:AAS}
\end{figure}

To reduce the computational cost, the proposed AAS module can automatically identify the importance of each patch as shown in \figref{AAS}. 
% Due to only the class token will be sent to the classifier for predicting labels, we argue that the more a patch is relevant to the class patch, the more important it is.
We argue that the more a patch is relevant to other patches across all heads, the more important it is. Thus, the importance score of patch $x_i$ in $l^{th}$ layer can be computed by:
\begin{equation}
  s^l(x_i)=\frac{1}{N_h}\frac{1}{n}\sum^{N_h}_{h=1}\sum^{n}_{j=1}\text{Attn}^l_h(x_i,x_j),
  \label{equ:p_score}
\end{equation}
where the $N_h$ and $n$ denote the number of heads and patches, respectively.
Furthermore, according to statistical characteristics of importance score, we define a quantization strategy to finally allocate the bit-width for each patch without any hyperparameter:
\begin{equation}
  \text{Bit-width}(x_i)= 
\begin{cases}
  (k+2)\text{bit} & \text{if} s^l(x_i)>\theta_1, \\ 
  k\text{bit} & \text{otherwise}, \\ 
  (k-2)\text{bit} & \text{if} s^l(x_i)<\theta_2, 
  \label{equ:p_bit}
\end{cases}
\end{equation}
where the $k$ is allocated from GSR, and also denotes the average bit-width of $x_i$. Given the mean $m^l(x_i)$ and standard deviation $v^l(x_i)$ of $s^l(x_i)$, the $\theta_1$ and $\theta_2$ can be computed by:
\begin{align}
  &\theta_1=m^l(x_i)+v^l(x_i), \\
  &\theta_2=m^l(x_i)-v^l(x_i),
\end{align}
where the $m^l(x_i)$ and $v^l(x_i)$ of importance score is a measure of the quality and discrimination of each patch. 
For those important patches, a high $m^l(x_i)$ indicates they are the high-quality candidates, and a high $v^l(x_i)$ means they are the most discriminative patch. 
Using the mean $m^l(x_i)$ and standard deviation $v^l(x_i)$ as the thresholds, we can adaptively identify the importance of patches by their statistical characteristics.

\begin{figure}[t!]
  \centering
  \includegraphics[width=\linewidth]{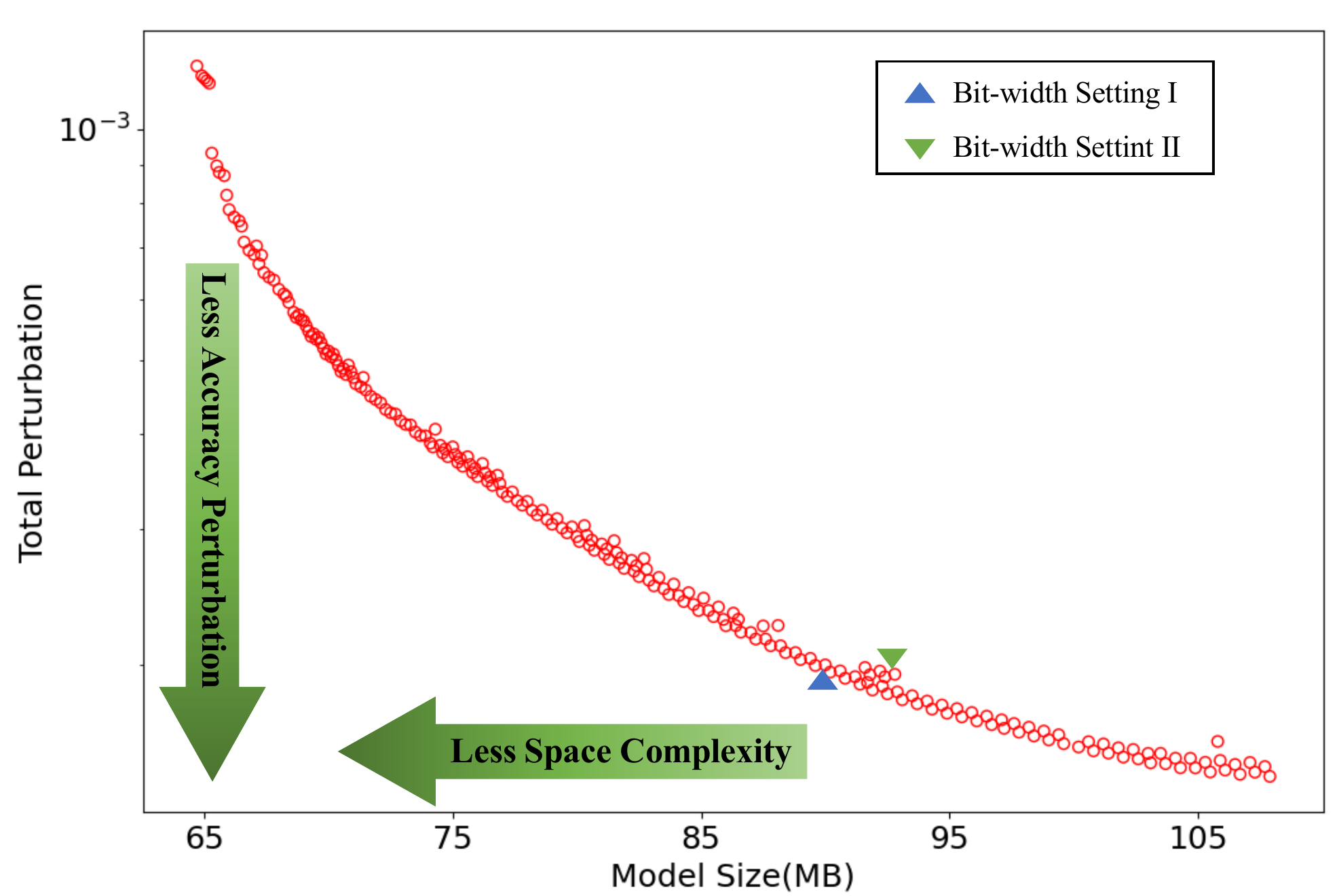}
  \vspace{-0.5cm}
  \caption{The Pareto Frontiers of ViT-Base\cite{vit}.  Here, each red point in the figure stands for a specific bit-precision setting, and the triangles represent the optimal trade-off points that are determined automatically.}
  \label{fig:Pareto}
\end{figure}

\subsection{Patch-wise Mixed Precision Quantization}
Finally, we design a search method to allocate bit-widths for each layer according to the proposed GSM automatically.

% Given the constraint $C$ and importance score $score_{i}$ of $i_{th}$ layer, we can reformulate the \equref{task} as:
% \begin{equation}
%   \min _{\mathcal{Q}_k(\mathbf{W})} \mathcal{L}(\mathcal{D}, \mathcal{Q}_k(\mathbf{W})) \quad \text { s.t. } \operatorname{Cost}(\mathrm{ \mathcal{Q}_k(\mathbf{W})}) \leq C,
% \end{equation}
Inspired by \cite{HAWQV2}, we use a Pareto frontier approach to solve \equref{task}. As shown in \figref{Pareto}, the main idea is to sort each candidate bit-width configuration based on the total second-order perturbation that they cause, according to the following metric:
\begin{equation}
  \Omega=\sum_{i=1}^L \Omega_i =\sum_{i=1}^L score_i \cdot\left\|\mathcal{Q}_k\left(\textbf{W}\right)-\textbf{W}\right\|_2^2,
  \label{equ:perturbation}
\end{equation}
where $i$ and $L$ refer to the $i_{th}$ component and the total number of components in the model, respectively. The $score_i$ is the importance score of $i_{th}$ layer, and $\left\|\mathcal{Q}_k\left(\textbf{W}\right)-\textbf{W}\right\|_2^2$ is the L2 norm of quantization perturbation. 
We compute model size as the $\operatorname{Cost}(\cdot)$ in \equref{task}:
\begin{equation}
  M=\sum_{i=1}^L M_i=\sum_{i=1}^L n_i b_i.
  \label{equ:model_size}
\end{equation}
where the $n_i$ and $b_i$ denotes the total number and bit-width of weight parameters in $i_{th}$ layer.
The pipeline of PMQ is summarized in \algref{PMQ}. Note that, the proposed PMQ can be used as a plug-and-play module to combine with quantization-aware training or post-training quantization schemes. In this work, we utilize the PTQ as the quantization scheme.

\begin{algorithm}[!t]
  \small
  \caption{Patch-wise Mixed Precision Quantization of Vision Transformer}
  \label{alg:PMQ}
  \begin{algorithmic}[1]
  \REQUIRE ~~\\
  Full-precision model with parameters $\textbf{W}$ and layers $L$, calibration dataset $X$ satisfied i.i.d assumption.
  \ENSURE ~~\\
  \textbf{b} is the optimal bit allocation of the model.
  \vspace{2mm}
  \STATE \# Step1: Calculate sensitivity; 
  \STATE Input $X$ to ViTs;
  \FOR{$i=1,2,\cdots,L$}
  \STATE Compute importance score $Score_i$ based on \equref{score}
  \ENDFOR
  \STATE \# Step2: Calculate Pareto Frontier;
  \STATE Divide the model into $L/\alpha$ groups.
  \STATE Divide the x-axis (model size) of the Pareto Frontier into $\beta$ intervals.
  \STATE Initial bit-width candidate sets for weight and activation.
  \FOR{$j=1,2,\cdots,L/\alpha$}
  \FOR{$k=1,2,\cdots,\alpha$}
  \STATE Compute accuracy perturbation $\Omega_i$ based on \equref{perturbation}
  \STATE Compute model size $M_i$ based on \equref{model_size}
  \ENDFOR
  \FOR{$k=1,2,\cdots,\beta$}
  \STATE Sort the corresponding $\Omega_i$ in descending order and select top $k$ allocation $\textbf{b}$ that have the lowest perturbation.
  \ENDFOR
  \ENDFOR
  \STATE \# Step3: Quantize the Patches in each MSA;
  \FOR{$l=1,2,\cdots,L$}
  \FOR{$p=1,2,\cdots,n$}
  \STATE Compute importance score $s^l(x_p)$ based on \equref{p_score}.
  \ENDFOR
  \STATE Reallocate bit-width for patches based on \equref{p_bit}
  \ENDFOR 
  \STATE Perform Post-training Quantization.
  \end{algorithmic}
\end{algorithm}

\section{Experiments}
In this section, we construct extensive experiments on ImageNet \cite{imageNet} to demonstrate the effectiveness of the proposed PMQ.
We first compare PMQ with previous state-of-the-art methods on different vision transformer structures. 
Moreover, we also perform ablation studies to verify the importance of each component in the proposed method.
\subsection{Implementation Details}
We adopt the Pytorch and TIMM library to implement the proposed PMQ and conduct all experiments on ImageNet, a
standard image classification dataset with 1000 classes and an
$224 \times 224$ image size. Specifically, we random sample 32 images from ImageNet training set as the calibration set, and evaluate the effectiveness of PMQ using Deit and ViT series on the validation set.

Unless specific, we adopt symmetric and asymmetric quantization for weight and activation, respectively, and the default calibration strategy is percentile.
In particular, all the weights and activation in vision transform are quantized, and the bit-width is allocated in the 2-8 bit search space.
The proposed PMQ only needs a piece of Nvidia A6000 GPU and a single AMD EPYC 7272 12-Core Processor.

\subsection{Ablation Studies}
 We first explore the impact of different components. As shown in \tabref{componets}, we set Deit-S \cite{deit} with the quantizer `Percentile' as the baseline. Compared with the baseline, PMQ improves the performance by $\sim$0.69. 
To be more specific, when the GSM is applied, the ACC increases by 0.38, and the AAS can also enhance the score by 0.48 AP enhancement.
This result validates the effectiveness of the proposed GSM and AAS.

 As reported in \tabref{metric}, we compare the compute time between our method and Hessian-based metric. Our lightweight GSM is 8 times faster than Hessian-based metrics and more accurate with comparable perturbations, which is essential in the quantization of ViTs.
\begin{table}[t]
  \small
  \centering
  \caption{Ablation study of GSM and AAS on DeiT-S.}
  \begin{tabular}{ccccc}
  \toprule
  Model & bit-width & GSM & AAS & ACC (Top-1)\\
  \midrule
  \multirow{4}{*}{DeiT-S} & \multirow{4}{*}{8MP} & $\times$ & $\times$ & 75.08 \\
   & & \checkmark & $\times$ & 75.46 \\
   & & $\times$ & \checkmark & 75.56 \\
   & & \checkmark & \checkmark & 75.77\\
  \bottomrule
  \end{tabular}
  \label{tab:componets}
\end{table}
 
\begin{table}[t]
  \small
  \centering
  \caption{Ablation study of different sensitivity metrics. “Pert.” means perturbation.}
  \begin{tabular}{ccccc}
  \toprule
  Metric & W/A/P  & Time (ms) & Pert. & Acc(Top-1)\\
  \midrule
  Baseline & 32/32/32 & - & 0 & 79.82\\
  Hessian &8MP/8MP/4MP & 68.8 & 0.32 & 74.98 \\
  GSM(ours) & 8MP/8MP/4MP & 7.94 & 0.37 & 75.77 \\
  \bottomrule
  \end{tabular}
  \label{tab:metric}
\end{table}

\begin{table}[htp]
  \small
  \centering
  \caption{Comparison of the quantization results on ImageNet dataset. `MP’ represents mixed-precision. `W/A/P' represents the bit widths for weight, activation, and patch in models. We also quantize the first and last layers.}
  \begin{tabular}{@{}ccccc@{}}
  \toprule
    Model & Method & W/A/P & Size (MB) & Acc (Top-1)\\
  \midrule
  \multirow{10}{*}{DeiT-S} & Baseline & 32/32/32 & 88 & 79.82 \\
  % \cline{2-8}
  %  & EMA & 8 & 8 & 8 & 22 &  & 75.962 \\
  %  & Percentile & 8 & 8 & 8 & 22 &  & 76.754 \\
  %  & OMSE & 8 & 8 & 8 & 22 &  &  \\
  %  & PMQ & 8MP & 8MP & 8MP & 21.1 &  & 76.886 \\
  \cline{2-5}
  %  & EMA & 8/8/4 & 22 & 73.58 \\
   & Percentile & 8/8/4 & 22 & 75.08(-4.74) \\
   & OMSE & 8/8/4 & 22 & 74.44(-5.38) \\
   & Liu$^*$\cite{ptqforvit} & 8/8/8(MP) & 22.2 & 78.09(-1.73)\\
   & PMQ(ours) & 8/8/4(MP) & 21.2 & \textbf{75.77}(-4.05)\\

   \cline{2-5}
  %  & EMA & 4/8/8 & 11 & 72.242 \\
   & Percentile & 4/8/8 & 11 & 72.69(-7.13) \\
   & OMSE & 4/8/8  & 11 & 70.71(-9.7) \\
   & PMQ(ours) & 4/8/8(MP) & 11.2 & \textbf{73.69}(-6.13) \\
   \cline{2-5}
  %  & EMA & 6/6/6 & 16.5 & 0.156 \\
   & Percentile & 6/8/6 & 16.5 & 76.03(-3.79) \\
   & OMSE & 6/8/6 & 16.5 & 74.84(-4.98) \\
   & Liu$^*$\cite{ptqforvit} & 6/6/6(MP) & 16.6 & 75.10(-4.72)\\
   & PMQ(ours) & 6/6/6(MP) & 16.5 & \textbf{76.68}(-3.14) \\
   \midrule
  \multirow{10}{*}{DeiT-B} & Baseline & 32/32/32 & 344 & 81.80 \\
  % \cline{2-8}
  %  & EMA & 8 & 8 & 8 & 22 &  & 75.962 \\
  %  & Percentile & 8 & 8 & 8 & 22 &  & 76.754 \\
  %  & OMSE & 8 & 8 & 8 & 22 &  &  \\
  %  & PMQ & 8MP & 8MP & 8MP & 21.1 &  & 76.886 \\
   \cline{2-5}
  %  & EMA & 8 & 8 & 4 &  &  & 73.582 \\
   & Percentile & 8/8/4 & 86.0 & 78.00(-3.8) \\
   & OMSE & 8/8/4 & 86.0 & 79.57(-2.23) \\
   & Liu$^*$\cite{ptqforvit} & 8/8/8(MP) & 86.8 & 81.29(-0.51)\\
   & PMQ(ours) & 8/8/4(MP) & 86.0 & \textbf{79.77}(-2.03) \\
   \cline{2-5}
  %  & EMA &  &  &  &  &  &  \\
   & Percentile & 4/8/8 & 43 & 76.66(-5.14) \\
   & OMSE & 4/8/8 & 43 & 78.12(-3.68) \\
   & PMQ(ours) & 4/8/8(MP) & 43.4 & \textbf{78.85}(-2.95) \\
   \cline{2-5}
  %  & EMA &  &  &  &  &  &  \\
   & Percentile & 6/8/6 & 64.5 & 78.44(-3.36) \\
   & OMSE & 6/8/6 & 64.5 & 79.62(-2.18) \\
   & Liu$^*$\cite{ptqforvit} & 6/6/6(MP) & 64.3 & 77.47(-4.33)\\
   & PMQ(ours) & 6/6/6(MP) & 64.5 & \textbf{79.64}(-2.16) \\
  \midrule
  \multirow{10}{*}{ViT-B} & Baseline & 32/32/32 & 344 & 84.53 \\
  % \cline{2-8}
  %  & EMA & 8 & 8 & 8 & 22 &  & 75.962 \\
  %  & Percentile & 8 & 8 & 8 & 22 &  & 76.754 \\
  %  & OMSE & 8 & 8 & 8 & 22 &  &  \\
  %  & PMQ & 8MP & 8MP & 8MP & 21.1 &  & 76.886 \\
   \cline{2-5}
  %  & EMA & 8 & 8 & 4 &  &  & 73.582 \\
   & Percentile & 8/8/4 & 86.2 & 48.60(-35.93) \\
   & OMSE & 8/8/4 & 86.2 & 73.04(-11.49) \\
   & Liu$^*$\cite{ptqforvit} & 8/8/8(MP) & 86.5 & 76.98(-7.55)\\
   & PMQ(ours) & 8/8/4(MP) & 85.5 & \textbf{73.46}(-11.07) \\
   \cline{2-5}
  %  & EMA &  &  &  &  &  &  \\
   & Percentile & 4/8/8& 43 & 32.15(-52.38) \\
   & OMSE & 4/8/8 & 43 & 66.16(-18.37) \\
   & PMQ(ours) & 4/8/8(MP) & 43.2 & \textbf{68.91}(-15.62) \\
   \cline{2-5}
  %  & EMA & &  &  &  &  &  \\
   & Percentile & 6/8/6 & 64.5 & 50.25(-34.28) \\
   & OMSE & 6/8/6 & 64.5 & 73.20(-11.33) \\
   & Liu$^*$\cite{ptqforvit} & 6/6/6(MP) & 64.8 & 75.26(-9.27)\\
   & PMQ(ours) & 6/6/6(MP) & 64.5 & \textbf{73.33}(-11.20)\\
  \bottomrule
  \multicolumn{5}{l}{$^*$results are from papers}%, and this method include a custom Post-training Quantization method.}
  \end{tabular}
  \label{tab:sota}
\end{table}
\subsection{Comparison with State-of-the-art Methods}
To verify the effectiveness of PMQ, we conduct extensive experiments on ImageNet with various vision transformers.
We employ several current post-training quantization methods in this paper, such as Percentile and OMSE.
We use the bit allocation algorithm with PMQ and combine it with the Percentile quantization scheme as our result.
Note that Liu\cite{ptqforvit} introduces a customized post-training quantization method to obtain better accuracy. Despite the above, our results are still competitive, especially after low-bit quantization of the patches.
As shown in \tabref{sota}, it can be observed that PMQ outperforms other unified quantization methods under different model constraints.
Specifically, when we quantize the weights, activations, and patches to 8 bit, 8 bit, and 4 bit, PMQ outperforms the Percentile by 1.77 on Deit-B, which demonstrates the effectiveness of our AAS.  
Moreover, PMQ also surpasses the Percentile and Liu\cite{ptqforvit} with the bit config 6/6/6(MP) on Deit-S.
There are similar results on DeiT-B and ViT-B. 
These results clearly evidence the advantage of the proposed PMQ in the ViT models.

\section{Conclusion}
In this work, we proposed a patch-wise mixed precision quantization, dubbed PMQ, for efficient inference of vision transformers.
To reduce the time consumption of sensitivity ranking, we design a lightweight criterion GSM, which can speed up the ranking process with comparable performance.
We also introduce a Pareto frontier method to automatically assign the optimal bit-width for each component of ViTs.
Moreover, we argue that not all patches have equal importance and thus design the AAS to further reduce the computational complexity of MSA.
Extensive experiments with different ViT models on ImageNet demonstrate the effectiveness of PMQ.
In the future, we will study the PMQ combined with QAT to improve the accuracy and use more constraints, including BitOPS and specific hardware, to find the optimal bit-width.

\section*{Acknowledgment}
This work was supported in part by the Scientific Instrument Developing Project of the Chinese Academy of Sciences under Grant YJKYYQ20200045; in part by the National Natural Science Foundation of China under Grant 62276255.

\bibliographystyle{IEEEtran}
\bibliography{conference}

\begin{thebibliography}{10}
\providecommand{\url}[1]{#1}
\csname url@samestyle\endcsname
\providecommand{\newblock}{\relax}
\providecommand{\bibinfo}[2]{#2}
\providecommand{\BIBentrySTDinterwordspacing}{\spaceskip=0pt\relax}
\providecommand{\BIBentryALTinterwordstretchfactor}{4}
\providecommand{\BIBentryALTinterwordspacing}{\spaceskip=\fontdimen2\font plus
\BIBentryALTinterwordstretchfactor\fontdimen3\font minus
  \fontdimen4\font\relax}
\providecommand{\BIBforeignlanguage}[2]{{%
\expandafter\ifx\csname l@#1\endcsname\relax
\typeout{** WARNING: IEEEtran.bst: No hyphenation pattern has been}%
\typeout{** loaded for the language `#1'. Using the pattern for}%
\typeout{** the default language instead.}%
\else
\language=\csname l@#1\endcsname
\fi
#2}}
\providecommand{\BIBdecl}{\relax}
\BIBdecl

\bibitem{vit}
A.~Dosovitskiy, L.~Beyer, A.~Kolesnikov, D.~Weissenborn, X.~Zhai,
  T.~Unterthiner, M.~Dehghani, M.~Minderer, G.~Heigold, S.~Gelly, J.~Uszkoreit,
  and N.~Houlsby, ``An image is worth 16$\times$16 words: Transformers for
  image recognition at scale,'' in \emph{ICLR}, 2021.

\bibitem{deit}
H.~Touvron, M.~Cord, M.~Douze, F.~Massa, A.~Sablayrolles, and H.~J{\'e}gou,
  ``Training data-efficient image transformers \& distillation through
  attention,'' in \emph{ICML}, 2021.

\bibitem{detr}
N.~Carion, F.~Massa, G.~Synnaeve, N.~Usunier, A.~Kirillov, and S.~Zagoruyko,
  ``End-to-end object detection with transformers,'' in \emph{ECCV}, 2020.

\bibitem{swin}
Z.~Liu, Y.~Lin, Y.~Cao, H.~Hu, Y.~Wei, Z.~Zhang, S.~Lin, and B.~Guo, ``Swin
  transformer: Hierarchical vision transformer using shifted windows,'' in
  \emph{ICCV}, 2021.

\bibitem{setr}
S.~Zheng, J.~Lu, H.~Zhao, X.~Zhu, Z.~Luo, Y.~Wang, Y.~Fu, J.~Feng, T.~Xiang,
  P.~H. Torr \emph{et~al.}, ``Rethinking semantic segmentation from a
  sequence-to-sequence perspective with transformers,'' in \emph{CVPR}, 2021.

\bibitem{FixedPoint}
D.~D. Lin, S.~S. Talathi, and V.~S. Annapureddy, ``Fixed point quantization of
  deep convolutional networks,'' in \emph{ICML}, 2016.

\bibitem{Fixed-point}
S.~Shin, K.~Hwang, and W.~Sung, ``Fixed-point performance analysis of recurrent
  neural networks,'' in \emph{ICASSP}, 2016.

\bibitem{Degree-Quant}
S.~A. Tailor, J.~Fern{\'{a}}ndez{-}Marqu{\'{e}}s, and N.~D. Lane,
  ``Degree-quant: Quantization-aware training for graph neural networks,'' in
  \emph{ICLR}, 2021.

\bibitem{QAT2}
P.~Stock, A.~Fan, B.~Graham, E.~Grave, R.~Gribonval, H.~J{\'{e}}gou, and
  A.~Joulin, ``Training with quantization noise for extreme model
  compression,'' in \emph{ICLR}, 2021.

\bibitem{QAT1}
S.~Zhou, Z.~Ni, X.~Zhou, H.~Wen, Y.~Wu, and Y.~Zou, ``Dorefa-net: Training low
  bitwidth convolutional neural networks with low bitwidth gradients,''
  \emph{CoRR}, vol. abs/1606.06160, 2016.

\bibitem{QAT3}
B.~Zhuang, L.~Liu, M.~Tan, C.~Shen, and I.~D. Reid, ``Training quantized neural
  networks with a full-precision auxiliary module,'' in \emph{CVPR}, 2020.

\bibitem{QAT4}
Z.~G. Liu and M.~Mattina, ``Learning low-precision neural networks without
  straight-through estimator {(STE)},'' in \emph{IJCAI}, S.~Kraus, Ed., 2019.

\bibitem{BRECQ}
Y.~Li, R.~Gong, X.~Tan, Y.~Yang, P.~Hu, Q.~Zhang, F.~Yu, W.~Wang, and S.~Gu,
  ``{BRECQ:} pushing the limit of post-training quantization by block
  reconstruction,'' in \emph{ICLR}, 2021.

\bibitem{Near-Lossless}
J.~Fang, A.~Shafiee, H.~Abdel{-}Aziz, D.~Thorsley, G.~Georgiadis, and
  J.~Hassoun, ``Near-lossless post-training quantization of deep neural
  networks via a piecewise linear approximation,'' \emph{arXiv preprint
  arXiv:2002.00104}, 2020.

\bibitem{PTQ1}
R.~Banner, Y.~Nahshan, and D.~Soudry, ``Post training 4-bit quantization of
  convolutional networks for rapid-deployment,'' in \emph{NeurIPS}, H.~M.
  Wallach, H.~Larochelle, A.~Beygelzimer, F.~d'Alch{\'{e}}{-}Buc, E.~B. Fox,
  and R.~Garnett, Eds., 2019.

\bibitem{PTQ2}
M.~Nagel, R.~A. Amjad, M.~van Baalen, C.~Louizos, and T.~Blankevoort, ``Up or
  down? adaptive rounding for post-training quantization,'' in \emph{ICML},
  2020.

\bibitem{PTQ3}
I.~Hubara, Y.~Nahshan, Y.~Hanani, R.~Banner, and D.~Soudry, ``Improving post
  training neural quantization: Layer-wise calibration and integer
  programming,'' \emph{CoRR}, 2020.

\bibitem{HAWQV1}
Z.~Dong, Z.~Yao, A.~Gholami, M.~W. Mahoney, and K.~Keutzer, ``{HAWQ:} hessian
  aware quantization of neural networks with mixed-precision,'' in \emph{ICCV},
  2019.

\bibitem{HAWQV2}
Z.~Dong, Z.~Yao, D.~Arfeen, A.~Gholami, M.~W. Mahoney, and K.~Keutzer,
  ``{HAWQ-V2:} hessian aware trace-weighted quantization of neural networks,''
  in \emph{NeurIPS}, 2020.

\bibitem{HAWQV3}
Z.~Yao, Z.~Dong, Z.~Zheng, A.~Gholami, J.~Yu, E.~Tan, L.~Wang, Q.~Huang,
  Y.~Wang, M.~W. Mahoney, and K.~Keutzer, ``{HAWQ-V3:} dyadic neural network
  quantization,'' in \emph{ICML}, 2021.

\bibitem{OMPQ}
Y.~Ma, T.~Jin, X.~Zheng, Y.~Wang, H.~Li, G.~Jiang, W.~Zhang, and R.~Ji,
  ``{OMPQ:} orthogonal mixed precision quantization,'' \emph{arXiv preprint
  arXiv:2109.07865}, 2021.

\bibitem{SAQ}
J.~Liu, J.~Cai, and B.~Zhuang, ``Sharpness-aware quantization for deep neural
  networks,'' \emph{arXiv preprint arXiv:2111.12273}, 2021.

\bibitem{HAQ}
K.~Wang, Z.~Liu, Y.~Lin, J.~Lin, and S.~Han, ``{HAQ:} hardware-aware automated
  quantization with mixed precision,'' in \emph{CVPR}, 2019.

\bibitem{AutoQ}
Q.~Lou, F.~Guo, M.~Kim, L.~Liu, and L.~Jiang, ``Autoq: Automated kernel-wise
  neural network quantization,'' in \emph{ICLR}, 2020.

\bibitem{DNAS}
B.~Wu, Y.~Wang, P.~Zhang, Y.~Tian, P.~Vajda, and K.~Keutzer, ``Mixed precision
  quantization of convnets via differentiable neural architecture search,''
  \emph{arXiv preprint arXiv:1812.00090}, 2018.

\bibitem{DQ}
S.~Uhlich, L.~Mauch, F.~Cardinaux, K.~Yoshiyama, J.~A. Garc{\'{\i}}a,
  S.~Tiedemann, T.~Kemp, and A.~Nakamura, ``Mixed precision dnns: All you need
  is a good parametrization,'' in \emph{ICLR}, 2020.

\bibitem{mobilevit}
S.~Mehta and M.~Rastegari, ``Mobilevit: Light-weight, general-purpose, and
  mobile-friendly vision transformer,'' in \emph{ICLR}, 2022.

\bibitem{mobilenetv2}
M.~Sandler, A.~G. Howard, M.~Zhu, A.~Zhmoginov, and L.~Chen, ``Mobilenetv2:
  Inverted residuals and linear bottlenecks,'' in \emph{CVPR}, 2018.

\bibitem{minivit}
J.~Zhang, H.~Peng, K.~Wu, M.~Liu, B.~Xiao, J.~Fu, and L.~Yuan, ``Minivit:
  Compressing vision transformers with weight multiplexing,'' in \emph{CVPR},
  2022.

\bibitem{evovit}
Y.~Xu, Z.~Zhang, M.~Zhang, K.~Sheng, K.~Li, W.~Dong, L.~Zhang, C.~Xu, and
  X.~Sun, ``Evo-vit: Slow-fast token evolution for dynamic vision
  transformer,'' in \emph{AAAI}, 2022.

\bibitem{levit}
B.~Graham, A.~El{-}Nouby, H.~Touvron, P.~Stock, A.~Joulin, H.~J{\'{e}}gou, and
  M.~Douze, ``Levit: a vision transformer in convnet's clothing for faster
  inference,'' in \emph{ICCV}, 2021.

\bibitem{FIM1}
P.~Molchanov, A.~Mallya, S.~Tyree, I.~Frosio, and J.~Kautz, ``Importance
  estimation for neural network pruning,'' in \emph{CVPR}, 2019.

\bibitem{FIM2}
L.~Theis, I.~Korshunova, A.~Tejani, and F.~Husz{\'a}r, ``Faster gaze prediction
  with dense networks and fisher pruning,'' \emph{arXiv preprint
  arXiv:1801.05787}, 2018.

\bibitem{imageNet}
J.~Deng, W.~Dong, R.~Socher, L.~Li, K.~Li, and L.~Fei{-}Fei, ``Imagenet: {A}
  large-scale hierarchical image database,'' in \emph{CVPR}, 2009.

\bibitem{ptqforvit}
Z.~Liu, Y.~Wang, K.~Han, W.~Zhang, S.~Ma, and W.~Gao, ``Post-training
  quantization for vision transformer,'' 2021.

\end{thebibliography}

\end{document}